\documentclass[journal,twoside,web]{ieeecolor}
\usepackage{generic}

\usepackage{inputenc}
\usepackage[T1]{fontenc}
\usepackage{lmodern}

\usepackage{setspace}
\usepackage{layout}

\usepackage[backend=biber,
    bibencoding=utf8,
    language=auto,
    style=ieee,
    sorting=none,
    maxbibnames=3, 
    minbibnames=3,
    maxcitenames=3,
    mincitenames=1,
    natbib=false]{biblatex}

\usepackage{amssymb}
\usepackage[fixamsmath]{mathtools}
\usepackage[version=4]{mhchem}
\usepackage{stmaryrd}
\DeclarePairedDelimiter\parentheses{\lparen}{\rparen}
\DeclarePairedDelimiter\abs{\lvert}{\rvert}
\DeclarePairedDelimiter\braces{\lbrace}{\rbrace}
\DeclarePairedDelimiter\brackets{\lbrack}{\rbrack}
\DeclarePairedDelimiter\bbrackets{\llbracket}{\rrbracket}
\newcommand{\func}[2]{\ensuremath{#1 \parentheses*{ #2 }}}

\usepackage{booktabs}
\usepackage[flushleft]{threeparttable}
\usepackage{caption}
\usepackage{subcaption}
\usepackage{multirow}
\usepackage{array}
\usepackage{tabularx}
\newcolumntype{Y}{>{\centering\arraybackslash}X}

\newcolumntype{L}{>{$}l<{$}} 
\newcolumntype{C}{>{$}l<{$}} 

\usepackage{acronym}
\usepackage[usenames,svgnames]{xcolor}
\usepackage[inline]{enumitem}
\usepackage[colorlinks=false]{hyperref}
\usepackage[capitalize]{cleveref}
\usepackage{microtype}
\microtypecontext{spacing=french}
\usepackage{nth}
\usepackage{siunitx}
\usepackage{textgreek}
\usepackage{xspace}

\newcommand{\describe}[1]{\acs{#1}:~\acl{#1}}
\newcommand{\ie}{i.\,e.\xspace}

\newcommand{\eg}{e.\,g.\xspace}

\newcommand{\R}[1]{\mathbb{R}^{#1}\xspace}
\newcommand{\N}{\mathbb{N}}
\newcommand{\train}{\ensuremath{\mathcal{D}_\text{train}}\xspace}
\newcommand{\eval}{\ensuremath{\mathcal{D}_\text{eval}}\xspace}
\newcommand{\test}{\ensuremath{\mathcal{D}_\text{test}}\xspace}
\newcommand{\ntrain}{\ensuremath{n_\text{train}}\xspace}
\newcommand{\neval}{\ensuremath{n_\text{eval}}\xspace}
\newcommand{\ntest}{\ensuremath{n_\text{test}}\xspace}
\newcommand{\ci}[4]{\num[separate-uncertainty]{#1\pm#2} \ensuremath{[#3,#4]}\xspace}
\newcommand{\plusminus}[2]{\ensuremath{\num[separate-uncertainty]{#1 \pm #2}}\xspace}

\def\BibTeX{{\rm B\kern-.05em{\sc i\kern-.025em b}\kern-.08em
    T\kern-.1667em\lower.7ex\hbox{E}\kern-.125emX}}

\begin{document}
\title{MSED: A Multi-Modal Sleep Event Detection Model for Clinical Sleep Analysis}

\author{Alexander Neergaard Zahid, \IEEEmembership{Member, IEEE}, Poul Jennum, Emmanuel Mignot, and Helge B. D. Sorensen, \IEEEmembership{Senior Member, IEEE}
\thanks{A. N. Zahid is with the Department of Applied Mathematics and Computer Science, Technical University of Denmark, Kgs. Lyngby, Denmark (e-mail: aneol@dtu.dk).}
\thanks{H. B. D. Sorensen is with the Department of Health Technology, Technical University of Denmark, Kgs. Lyngby, Denmark.}
\thanks{E. Mignot is with the Stanford Center for Sleep Sciences and Medicine, Stanford University, Palo Alto, CA, USA.}
\thanks{P. Jennum is with the Danish Center for Sleep Medicine, University Hospital of Copenhagen, Glostrup, Denmark.}
\thanks{Copyright (c) 2023 IEEE. Personal use of this material is permitted. However, permission to use this material for any other purposes must be obtained from the IEEE by sending an email to \url{pubs-permissions@ieee.org}.}}

\maketitle

\acrodef{AASM}{American Academy of Sleep Medicine}
\acrodef{Adam}{adaptive moment estimation}
\acrodef{AHI}{apnea-hypopnea index}
\acrodef{ASDA}{American Sleep Disorders Association}
\acrodef{ANOVA}{analysis of variance}
\acrodef{Ar}{arousal}
\acrodef{ArI}{arousal index}
\acrodef{BF}{basal forebrain}
\acrodef{bGRU}{bidirectional gated recurrent unit}
\acrodef{BMI}{body-mass index}
\acrodef{BN}{batch normalization}
\acrodef{CNN}{convolutional neural network}
\acrodef{conv}{convolution}
\acrodef{CSA}{central sleep apnea}
\acrodef{CSF}{cerebrospinal fluid}
\acrodef{DMH}{dorsomedial hypothalamic nucleus}
\acrodef{DOSED}{Dreem One Shot Event Detector}
\acrodef{DRN}{dorsal raphe nucleus}
\acrodef{ECG}{electrocardiography}
\acrodef{EEG}{electroencephalography}
\acrodef{EMG}{electromyography}
\acrodef{EOG}{electrooculography}
\acrodef{GABA}{gamma-aminobutyric acid}
\acrodef{GP}{Gaussian process}
\acrodef{GRU}{gated recurrent unit}
\acrodef{HLA}{human leukocyte antigen}
\acrodef{hcrt}{hypocretin}
\acrodef{ICC}{intraclass correlation coefficient}
\acrodef{ICSD}{International Classification of Sleep Disorders}
\acrodef{IoU}{intersection over union}
\acrodef{ISRUC}{Something}
\acrodef{LAMF}{low amplitude, mixed frequency}
\acrodef{LC}{locus coeruleus}
\acrodef{LDT}{laterodorsal tegmental nucleus}
\acrodef{LHA}{lateral hypothalamic area}
\acrodef{LM}{leg movement}
\acrodef{LMI}{leg movement index}
\acrodef{LPT}{lateral pontine tegmentum}
\acrodef{LSTM}{long short-term memory}
\acrodef{MASSCv1}{multi-modal automatic sleep stage classification version 1}
\acrodef{MASSCv2}{multi-modal automatic sleep stage classification version 2}
\acrodef{MCH}{melanin-concentrating hormone}
\acrodef{MnPO}{median preoptic nucleus}
\acrodef{MrOS}{MrOS Sleep Study}
\acrodef{MSED}{multi-modal sleep event detection}
\acrodef{MSL}{mean sleep latency}
\acrodef{MSLT}{multiple sleep latency test}
\acrodef{N1}{non-\protect\ac{REM} stage 1}
\acrodef{N2}{non-\protect\ac{REM} stage 2}
\acrodef{N3}{non-\protect\ac{REM} stage 3}
\acrodef{NREM}{non-rapid eye movement}
\acrodef{NSRR}{National Sleep Research Resource}
\acrodef{NT1}{narcolepsy type 1}
\acrodef{NT2}{narcolepsy type 2}
\acrodef{OSA}{obstructive sleep apnea}
\acrodef{PB}{parabrachial nucleus}
\acrodef{PC}{preocoeruleus nucleus}
\acrodef{PGO}{ponto-geniculo-occipital}
\acrodef{PLM}{periodic leg movement}
\acrodef{PLMI}{periodic leg movement index}
\acrodef{PLMS}{periodic leg movement in sleep}
\acrodef{POA}{preoptic area}
\acrodef{PPT}{pedunculopontine tegmental nucleus}
\acrodef{PSG}{polysomnography}
\acrodef{RK}[R\&K]{Rechtschaffen and Kales}
\acrodef{RBD}{\protect\acs{REM} sleep behaviour disorder}
\acrodef{RDI}{respiratory disturbance index}
\acrodef{REM}{rapid eye movement}
\acrodef{REML}{\protect\acs{REM} sleep latency}
\acrodef{ReLU}{rectified linear unit}
\acrodef{RFE}{recursive feature elimination}
\acrodef{RNN}{recurrent neural network}
\acrodef{ROC}{receiver operating characteristic}
\acrodef{SDB}{sleep disordered breathing}
\acrodef{SCN}{suprachiasmatic nucleus}
\acrodef{SE}{sleep efficiency}
\acrodef{SEM}{slow eye movement}
\acrodef{SHHS}{Sleep Heart Health Study}
\acrodef{SL}{sleep latency}
\acrodef{SLD}{sublaterodorsal nucleus}
\acrodef{SOREMP}{sleep onset \protect\acs{REM} period}
\acrodef{SSC}{Stanford Sleep Cohort}
\acrodef{SWA}{slow wave activity}
\acrodef{SWS}{slow wave sleep}
\acrodef{TMN}{tuberomammillary nucleus}
\acrodef{TST}{total sleep time}
\acrodef{vlPAG}{ventrolateral periaqueductal gray}
\acrodef{VLPO}{ventrolateral preoptic nucleus}
\acrodef{vM}{ventral medulla}
\acrodef{vPAG}{ventral periaqueductal gray}
\acrodef{WASO}{wake after sleep onset}
\acrodef{WSC}{Wisconsin Sleep Cohort}
\acrodef{W}{wakefulness}

\acrodefplural{AHI}[AHIs]{apnea-hypopnea indices}
\acrodefplural{ArI}[ArIs]{arousal indices}
\acrodefplural{RDI}[RDIs]{respiratory disturbance indices}
\acrodefplural{PLMI}[PLMIs]{periodic leg movement indices}
\acrodefplural{PSG}[PSGs]{polysomnographies}

\begin{abstract}
\label{sec:abstract}
Clinical sleep analysis require manual analysis of sleep patterns for correct diagnosis of sleep disorders. 
However, several studies have shown significant variability in manual scoring of clinically relevant discrete sleep events, such as arousals, leg movements, and sleep disordered breathing (apneas and hypopneas).
We investigated whether an automatic method could be used for event detection and if a model trained on all events (joint model) performed better than corresponding event-specific models (single-event models).
We trained a deep neural network event detection model on 1653 individual recordings and tested the optimized model on 1000 separate hold-out recordings. 
F1 scores for the optimized joint detection model were 0.70, 0.63, and 0.62 for arousals, leg movements, and sleep disordered breathing, respectively, compared to 0.65, 0.61, and 0.60 for the optimized single-event models. 
Index values computed from detected events correlated positively with manual annotations (r\textsuperscript{2} = 0.73, r\textsuperscript{2} = 0.77, r\textsuperscript{2} = 0.78, respectively). 
We furthermore quantified model accuracy based on temporal difference metrics, which improved overall by using the joint model compared to single-event models.
Our automatic model jointly detects arousals, leg movements and sleep disordered breathing events with high correlation with human annotations.
Finally, we benchmark against previous state-of-the-art multi-event detection models and found an overall increase in F1 score with our proposed model despite a 97.5\% reduction in model size.
Source code for training and inference is available at \url{https://github.com/neergaard/msed.git}.
\end{abstract}


\section{Introduction}
\label{sec:introduction}
Clinical sleep analysis is currently evaluated manually by experts based on guidelines from the \ac{AASM} detailed in the \ac{AASM} Scoring Manual~\cite{Berry2020}.
The guidelines detail both technical and clinical best practices for setting up and recording \acp{PSG}, which are overnight recordings of various electrophysiological signals including \ac{EEG}, \ac{EOG}, chin and leg \ac{EMG}, \ac{ECG}, respiratory inductance plethysmography from the thorax and abdomen, oronasal pressure, and blood oxygen levels.
Based on these signals, expert technicians score and analyse the \ac{PSG} for sleep stages [\ac{W}, \ac{REM} sleep, \ac{N1}, \ac{N2}, and \ac{N3}], and sleep micro-events summarized by key metrics, such as the number of apneas and hypopneas per hour of sleep (\acl{AHI}, \acs{AHI}), the number of (periodic) leg movements per hour of sleep [(periodic) leg movement index, (P)LMI], and the number of arousals per hour of sleep (\acl{ArI}, \acs{ArI}).

\Acp{Ar} are defined as abrupt shifts in \ac{EEG} frequencies towards alpha, theta, and beta rhythms for at least \SI{3}{\second} with a preceding period of stable sleep of at least \SI{10}{\second}~\cite{Halasz2004}.
During \ac{REM} sleep, where the background \ac{EEG} shows similar rhythms, arousal scoring requires a concurrent increase in chin \ac{EMG} lasting at least \SI{1}{\second}~\cite{Berry2020}.

\Acp{LM} should be scored when there is an increase in amplitude of at least \SI{8}{\micro\volt} in the leg \ac{EMG} channels above baseline level with a duration between \SIrange{0.5}{10}{\second}~\cite{ferri_world_2016}.
A PLM series is then defined as a sequence of 4 \acp{LM}, where the time between \ac{LM} onsets is between \SIrange{5}{90}{\minute}~\cite{Berry2020, Ferri2017a}.
\begin{figure*}[tbp]
    \centering
    \includegraphics[width=\textwidth]{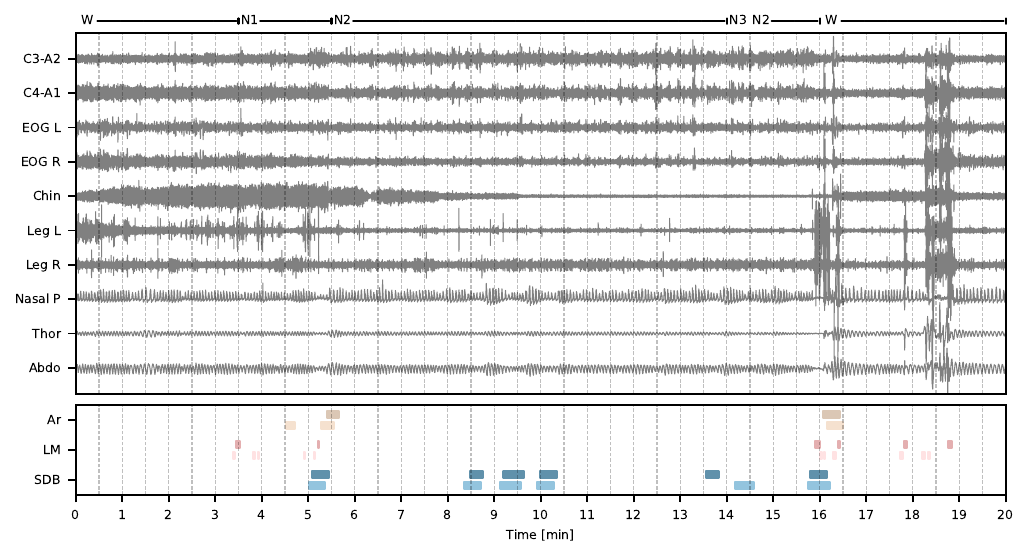}
    \caption{%
    Example of MSED input and output predictions. 
    Ten PSG channels comprising EEG, EOG, EMG, and breathing are visualized, which are fed to the MSED model.
    The bottom shows the manually scored (darker shade) and model predicted (lighter shade) events associated with the given recording segment.
    Sleep stages are shown at the top. %
    \describe{W}; %
    \describe{N1}; %
    \describe{N2}; %
    \describe{N3}; %
    \describe{Ar}; %
    \describe{LM}; %
    \describe{SDB}.
    }
    \label{fig:example}
\end{figure*}

Apneas are generally scored when there is a complete (\SI{>=90}{\percent} of pre-event baseline) cessation of breathing activity.
The underlying cause can be either a physical obstruction (obstructive apnea) or due to an underlying disruption in the central nervous system control (central apnea) for at least \SI{10}{\second}~\cite{Rosenberg2014}.
When breathing is only partially reduced (\SI{\geq30}{\percent} of pre-event baseline) and the duration of the excursion is \SI{>=10}{\second}, the event is scored as a hypopnea if there is either a \SI{>=4}{\percent} oxygen desaturation or a \SI{>=3}{\percent} oxygen desaturation coupled with an arousal~\cite{Berry2020}.
\Ac{SDB} here refers to the collective of apneas and hypopneas.

Several studies have shown significant variability in the scoring of both sleep stages~\cite{Norman2000, Danker-Hopfe2004, Danker-Hopfe2009, Rosenberg2013, Zhang2015a, Younes2016, Younes2018} and sleep micro-events~\cite{Drinnan1998, Whitney1998, Loredo1999, Smurra2001, Thomas2003, Bonnet2007, Magalang2013, Rosenberg2014}.
This has prompted extensive research into automatic methods for classifying sleep stages in large-scale studies~\cite{Koch2014, Supratak2017, Chambon2018c, Olesen2018c, Biswal2018, Stephansen2018, Phan2019a, Phan2019b}, while the research in automatic arousal~\cite{Olesen2019, Alvarez-Estevez2019, Brink-Kjaer2020} and \ac{LM}~\cite{Carvelli2020} detection on a similar scale is limited, but has attracted increased focus as evidenced by the \emph{You Snooze You Win} PhysioNet challenge from 2018~\cite{Ghassemi2018, Goldberger2000}. 
\citeauthor{Biswal2018} recently proposed a multi-task CNN/RNN combination model for the purpose of classifying sleep stages and predicting \ac{AHI} and \ac{LMI}~\cite{Biswal2018}.
They trained their model on \num{9000} \ac{PSG} recordings from the Massachusetts General Hospital (MGH) and evaluated their model on a held-out MGH dataset consisting of \num{1000} \acp{PSG}, and on \num{5804} \acp{PSG} studies from the \ac{SHHS}, yielding strong \ac{AHI}/expert correlation values (\num{0.85} on MGH, \num{0.77} on \ac{SHHS}) and \ac{LMI}/expert correlation (0.79 on MGH).
\citeauthor{Brink-Kjaer2020} published a CNN/RNN model for combined arousal and sleep/wake detection yielding an arousal detection F1 score of 0.79 on a test set of 1024 unique subjects~\cite{Brink-Kjaer2020}, which was subsequently validated in two separate patient groups~\cite{BrinkKjaer2021a, BrinkKjaer2021b}.
Similarly, \citeauthor{Carvelli2020} proposed a CNN/RNN model for \ac{LM} detection reporting an impressive F1 score of 0.77 on \num{348} PSGs from the MrOS sleep study~\cite{Carvelli2020}.
However, these models are all based on classifying windows of sleep data with subsequent manual fine-tuning and post-processing to combine events predicted in close-proximity windows, which incurs a human-factor bias.

Recently,~\citeauthor{Chambon2019} proposed the \ac{DOSED} algorithm for detecting sleep spindles and K-complexes in the sleep EEG~\cite{Chambon2019}, which was further developed for arousal and leg movement detection in subsequent publications by~\citeauthor{Olesen2019}~\cite{Olesen2019, Olesen2020b}.
The advantage of this kind of approach is two-fold: first, the model is trained end-to-end to detect and classify events of any type, since there is no reliance on manual post-hoc processing of event predictions; and second, using a grid of default event windows (discussed in \cref{sec:overview}) allows durations of different time scales.
However, as these models were only designed for either EEG-only events~\cite{Chambon2019,Olesen2020b}, or did not investigate joint detection of events occurring across multiple modalities~\cite{Olesen2019}, there is still an unmet need for models capable of predicting events of multiple classes from multiple sensor types.

In this study, we extend the previous work in~\cite{Chambon2019, Olesen2019, Olesen2020b} and introduce the \ac{MSED} model for joint detection of sleep micro-events.
The model combines multiple recording modalities from the \ac{PSG} and recent advances in machine learning to not only classify arousals, \acp{LM}, and \acp{SDB}, but also annotate them in the temporal domain without the need for post-hoc processing of predictions. 
An example of model predictions for a segment of PSG data is shown in~\cref{fig:example}, where the input signals are shown in the top box and manually scored/predicted events in the bottom box.

Our contributions are as follows:
\begin{itemize}
    \item We propose the MSED, which is a CNN/RNN based model using disentangled feature extraction streams trained end-to-end for for multi-modal sleep event detection. 
    To our knowledge, this is the first time this has been shown for multiple event types with multiple modalities trained simultaneously.
    \item We report inceased F1 scores using MSED compared to previous state-of-the-art in multi-event detection, despite a 97.5\% reduction in memory footprint as defined by the number of model parameters.
    \item Clinically relevant endpoints as computed by MSED correlate strongly with expert-scored values.
    \item Source code for training and testing models are available at \url{https://github.com/neergaard/msed.git}.
\end{itemize}

\section{Data}
\label{sec:data}
\begin{table*}[tb]
    \small
    \centering
    \renewcommand{\arraystretch}{1.2}
    \begin{threeparttable}
        \caption{MrOS demographics by subset.}\label{tab:demographics}
        \begin{tabularx}{\textwidth}{@{}lYYYc@{}}
        \toprule
        & \(\train\) & \(\eval\) & \(\test\) & \textit{p}-value \\
        \midrule
        \textit{n}                                  & 1653                          & 200                            & 1000                          & - \\
        Age, years                                  & \ci{76.4}{5.6}{67.0}{90.0}    & \ci{76.8}{5.4}{68.0}{90.0}     & \ci{76.4}{5.3}{67.0}{90.0}    & 0.404   \\
        \acs{BMI}, \si{\kilogram\per\square\meter} & \ci{27.3}{3.9}{16.0}{47.0}    & \ci{27.0}{3.6}{19.0}{40.0}     & \ci{27.0}{3.7}{17.0}{45.0}    & 0.247   \\ \midrule
        \acs{TST}, \si{\minute}                     & \ci{357.3}{69.0}{54.0}{615.0} & \ci{354.0}{69.1}{108.0}{503.0} & \ci{353.6}{68.7}{62.0}{572.0} & 0.312   \\
        \acs{SL}, \si{\minute}                      & \ci{22.9}{25.6}{1.0}{349.0}   & \ci{21.6}{23.0}{1.0}{135.0}    & \ci{25.1}{32.1}{1.0}{402.0}   & 0.284   \\
        \acs{REML}, \si{\minute}                    & \ci{109.5}{77.9}{0.0}{578.0}  & \ci{103.5}{70.0}{10.0}{413.0}  & \ci{107.2}{75.3}{3.0}{590.0}  & 0.466   \\
        \acs{WASO}, \si{\minute}                    & \ci{116.7}{67.1}{11.0}{462.0} & \ci{119.0}{70.8}{15.0}{372.0}  & \ci{112.9}{65.0}{6.0}{458.0}  & 0.471   \\
        \acs{SE}, \si{\percent}                     & \ci{75.9}{12.1}{17.0}{97.0}   & \ci{75.5}{12.3}{37.0}{96.0}    & \ci{76.4}{11.8}{26.0}{98.0}   & 0.690    \\
        \acs{N1}, \si{\percent}                     & \ci{6.8}{4.1}{0.0}{31.0}      & \ci{7.0}{4.5}{0.0}{28.0}       & \ci{6.9}{4.7}{1.0}{58.0}      & 0.968   \\
        \acs{N2}, \si{\percent}                     & \ci{62.7}{9.5}{28.0}{89.0}    & \ci{62.0}{9.7}{30.0}{90.0}     & \ci{62.8}{10.0}{21.0}{95.0}   & 0.451   \\
        \acs{N3}, \si{\percent}                     & \ci{11.4}{9.0}{0.0}{55.0} 	& \ci{11.8}{9.7}{0.0}{55.0} 	 & \ci{11.1}{9.0}{0.0}{57.0}     & 0.638 \\
        \acs{REM}, \si{\percent}                    & \ci{19.2}{6.5}{0.0}{44.0}     & \ci{19.4}{7.2}{0.0}{41.0}      & \ci{19.3}{6.7}{0.0}{42.0}     & 0.894   \\ \midrule
        \acs{ArI}, \si{\per\hour}                   & \ci{23.5}{11.8}{3.0}{87.0}    & \ci{23.4}{11.0}{4.0}{77.0}     & \ci{23.8}{11.8}{4.0}{102.0}   & 0.661   \\
        \acs{AHI}, \si{\per\hour}                   & \ci{13.5}{13.9}{0.0}{83.0}    & \ci{13.6}{13.3}{0.0}{59.0}     & \ci{14.2}{15.5}{0.0}{89.0}    & 0.907   \\
        \acs{PLMI}, \si{\per\hour}                  & \ci{35.4}{37.1}{0.0}{233.0}   & \ci{36.6}{39.0}{0.0}{178.0}    & \ci{36.0}{37.7}{0.0}{175.0}   & 0.993   \\ \bottomrule
        \end{tabularx}
        \begin{tablenotes}
            \item %
            \describe{BMI}; %
            \describe{TST}; %
            \describe{SL}; %
            \describe{REML}; %
            \describe{WASO}; %
            \describe{SE}; %
            \describe{N1}; %
            \describe{N2}; %
            \describe{N3}; %
            \describe{REM}; %
            \describe{ArI}; %
            \describe{AHI}; %
            \describe{PLMI}.
        \end{tablenotes}
    \end{threeparttable}
\end{table*}
We collected \acp{PSG} from the MrOS Sleep Study conducted between 2003 and 2005, an ancillary part of the larger Osteoporotic Fractures in Men Study~\cite{Blank2005,Orwoll2005,Blackwell2011}.
The main outcome of the MrOS Sleep Study was to investigate and discover connections between sleep disorders, skeletal fractures, and cardiovascular disease and mortality in community-dwelling older (\num{>65} years).
Of the original 5994 study participants, 3135 subjects were enrolled at one of six sites in the USA for a comprehensive sleep assessment, while 2909 of these underwent a first visit full-night in-home \ac{PSG} recording.
The \ac{PSG} studies were subsequently scored by certified sleep technicians according to the prevailing guidelines at the time.
Sleep stages were scored into stages 1, 2, 3, 4, wakefulness, and \ac{REM} according to \ac{RK} scoring guidelines. For the purpose of this study, sleep stages were converted into their \ac{AASM} equivalents (stage 1 into N1, stage 2 into N2, and stages 3 and 4 into N3)~\cite{Berry2020}.
Arousals were scored as abrupt increases in \ac{EEG} frequencies lasting at least \SI{3}{\second} according to older guidelines from the American Sleep Disorders Association~\cite{AmericanSleepDisordersAssociation1992}.
Apneas were defined as complete or near complete cessation of airflow lasting more than \SI{10}{\second} with an associated \SI{3}{\percent} or greater \ce{SaO2} desaturation, while hypopneas were based on a clear reduction in breathing of more than \SI{30}{\percent} deviation from baseline breathing lasting more than \SI{10}{\second}, and likewise assocated with a greater than \SI{3}{\percent} \ce{SaO2} desaturation.
While the scoring criteria for scoring \acp{LM} are not explicitly available for the MrOS Sleep Study, the prevailing standard at the time of the study was to score \acp{LM} following an increase in leg \ac{EMG} amplitude of more than \SI{8}{\micro\volt} above resting baseline levels for at least \SI{0.5}{\second}, but shorter than \SI{10}{\second}~\cite{Zucconi2006}.
A subset of the 2909 subjects also participated in follow-up sessions, although these studies did not include scoring of leg movements.

\subsection{Subset demographics and partitioning}
We used all first visit \ac{PSG} studies (\(n = 2853\)) available from the \ac{NSRR}~\cite{Dean2016,Zhang2018}, which we partitioned into a training set (\train, \( n_{\text{train}} = 1653 \)), a validation set (\eval, \(n_{\text{eval}} = 200\)), and a final testing set (\test, \(n_{\text{test}} = 1000 \)).
Key demographics and \ac{PSG}-related variables for each subset are shown as mean \(\pm\) standard deviation with range in parenthesis in~\cref{tab:demographics}.

\subsubsection{Signal and events}
For this study, we considered three \ac{PSG} events: \acp{Ar}, \acp{LM}, and \ac{SDB} events.
These types of events are based on a specific set of electrophysiological channels from the \ac{PSG} consisting of left and right central \ac{EEG} (C3 and C4), left and right \ac{EOG}, left and right chin \ac{EMG}, left and right leg \ac{EMG}, nasal pressure, and respiratory inductance plethysmography from the thorax and abdomen.
\Ac{EEG} and \ac{EOG} channels were referenced to the contralateral mastoid process, while a chin \ac{EMG} was synthesized by subtracting the right chin \ac{EMG} from the left chin \ac{EMG}.

Apart from the raw signal data, we also extracted onset time relative to the study start time and duration times for each event type in each \ac{PSG}.
    
\section{Methods}\label{sec:methods}
\subsection{Notation}
We denote by \(\llbracket a, b \rrbracket\) the set of integers \(\lbrace n \in \N \mid a \leq n \leq b\rbrace\) with \(\llbracket N \rrbracket\) being shorthand for \(\llbracket 1, N \rrbracket\), and by \(n \in \llbracket N \rrbracket \) the \textit{n}th sample in \(\llbracket N \rrbracket\).
A segment of PSG data is denoted by \(\mathbf{x} \in \R{C \times T}\), where \(C\) is the number of channels and \(T\) is the duration of the segment in samples.
An event type is defined as \(\varepsilon_i = \left( \varrho_i, \delta_i, l_i \right) \in \R{2}_{+} \times \mathcal{L}\), where \(\varrho,\delta,l\) is center point, duration, and label of the event, and \(\mathcal{L} = \bbrackets{L} \) is the event label space. The set of \(N_t\) true events for a given PSG segment is denoted by \( \boldsymbol{\varepsilon}^{t} = \braces*{ \varepsilon_{i}^{t} \mid i \in \bbrackets*{ N_{t} } } \). 
By \(\mathbf{\chi} \in \mathcal{D}_\ast\) we denote a sample in either one of the three subsets.
In the description of the network architecture, we have omitted the batch dimension from all calculations for brevity.
    
\subsection{Model overview}\label{sec:overview}
\begin{figure*}[tb]
    \centering
    \includegraphics[width=0.9\textwidth]{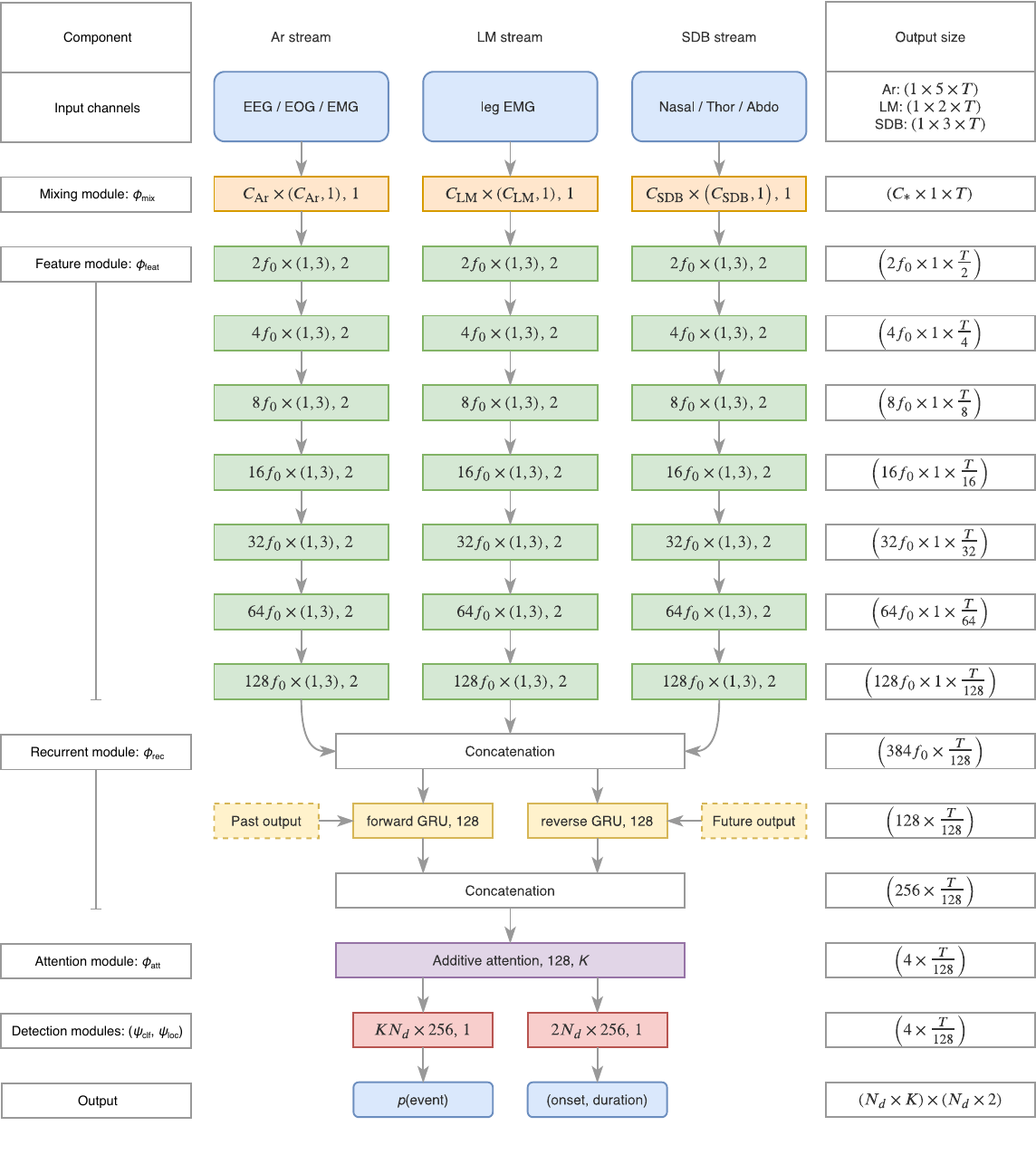}
    \caption{\acs{MSED} network architecture. 
    Left column contains component names, while right column shows the output dimensions for each operation as (number of filters[ x singleton] x time steps). 
    Each stream in the middle (green) processes a separate set of input channels (blue, top), the results of which are concatenated before the bGRU (yellow).
    Outputs from the additive attention layer (purple) are convolved in the final classification and localization layers (red) to output the probabilities for each event class, and the predicted onset and duration of each event (blue, bottom). 
    Convolution layers (orange, green, red) are detailed as [number of feature maps x kernel size, stride]. 
    Recurrent layer (yellow) shows the direction and number of hidden units. 
    Additive attention layer (purple) is described with the number of hidden and output units.
    \describe{GRU}; %
    \describe{Ar}; %
    \describe{LM}; %
    \describe{SDB}.
    }
    \label{fig:architecture}
\end{figure*}
Given an input set \( \boldsymbol{\chi} = \braces*{\mathbf{x}, \boldsymbol{\varepsilon}^t} \in \R{C \times T} \times \R{N_t \times 2}_+ \times \mathcal{L}\) containing \ac{PSG} data with \(C\) channels and \(T\) time steps, and true events \(\boldsymbol{\varepsilon}\), the goal of the model is to detect any possible events in the segment, where, in this context, detection covers both classification \textit{and} localization of any event in the segment space.

The model generates a set of \textit{default event windows} \(\boldsymbol{\varepsilon}^d = \braces*{ \varepsilon^{d}_{j} \mid j \in \bbrackets*{N_d} }\) for the current segment, and matches each true event to a default event window if their intersection-over-union (IoU) is at least 0.5.

At test time, we generate predictions across the default event windows and use a non-maximum suppression procedure to select between the candidate predictions.
For a given class \textit{k}, the procedure is as follows:
First, the predictions are sorted according to probability of the event, which is above a threshold \(\theta_k\).
Then, using the most probable prediction as an anchor, we sequentially evaluate the IoU between the anchor and the remaining candidate predictions, removing those with IoU larger than 0.5.

The output of the model is thus the set of predicted events \(\boldsymbol{\varepsilon}^{p} = \braces*{\mathbf{p}, \mathbf{y}}\) containing the predicted class probabilities along with the corresponding onsets and durations

\subsection{Signal conditioning}
We resampled all signals to a common sampling frequency of $f_s = \SI{128}{\hertz}$ using a poly-phase filtering approach (Kaiser window, \( \beta = \num{5.0} \)).
Based on recommended filter specifications from the~\ac{AASM}, we designed Butterworth IIR filters for four sets of signals~\cite{Berry2020}.
\Ac{EEG} and \ac{EOG} channels were filtered using a \nth{2} order filter with a \SIrange[range-phrase=--]{0.3}{35}{\hertz} passband, while chin and leg \ac{EMG} channels were filtered using a \nth{4} order high-pass filter with a \SI{10}{\hertz} cut-off frequency.
Nasal pressure channels was filtered using a \nth{4} order high-pass filter with a \SI{0.03}{\hertz} cut-off frequency, while thoracoabdominal channels were filtered using a \nth{2} order filter with a \SIrange[range-phrase=--]{0.1}{15}{\hertz} passband.
All filters were implemented using the zero-phase method.

Filtered signals were subsequently standardized by subtracting signal means and dividing by signal standard deviations for each \ac{PSG}.

\subsection{Target encoding}
For each data segment, target event classes \(\boldsymbol{\pi} \in \R{N_m \times K}\) were generated by one-hot encoding, and the target detection variable containing the onset and duration times \(\mathbf{t} \in \R{N_m \times 2}\) was encoded as
\begin{equation}
    t_i = \parentheses*{\frac{\varrho^m_i - \varrho^d_j}{\delta^d_j}, \, \log \frac{\delta^m_i}{\delta^d_j}}, \, i \in \bbrackets*{N_m}, \, j \in \bbrackets*{N_d},
\end{equation}
where \(\varrho^m_i\) is the center point of the true event matched to a default event window \(\varrho^d_j\), and \(\delta^m_i\) and \(\delta^d_j\) are the corresponding durations of the true and default events.

\subsection{Data sampling}
As the total number of default event windows \( N_d \) in a data segment most likely will be much higher than the number of event windows matched to a true event, \ie \( N_d \gg N_m\), we implemented a similar random data sampling strategy as in~\cite{Olesen2019}.
At training step \(t\), a given \ac{PSG} record \(r\) has a certain number of associated number of \ac{Ar}, \ac{LM}, and \ac{SDB} events (\(n_{\mathrm{Ar}},n_{\mathrm{LM}},n_{\mathrm{SDB}}\), respectively).
We randomly sample a class \(k\) with equal probability $p_k = \frac{1}{K - 1}$, while disregarding the negative class since this is most likely over-represented in the data segment in any case.
Given the class \(k\), we randomly sample an event \(\varepsilon_k\) from the PSG record \(r\).
We then randomly sample a \(C\times T\) data segment with start time in the range \( \left[ \bar{\varepsilon}_k - T, \bar{\varepsilon}_k \right]\) where \( \bar{\varepsilon}_k \) is the sample midpoint of the event \( \varepsilon_{k} \).
This ensures that a sampled data segment will contain at least \SI{50}{\percent} of at least one event.
We found that this approach to sampling data segments with a large ratio of negative to positive samples to be beneficial in all our experiments when monitoring the loss on the validation set.

\subsection{Network architecture}
Similar to the architecture described in~\cite{Brink-Kjaer2020}, we designed a splitstream network architecture, where each stream is responsible for the bulk feature extraction for a specific event class.
For the given problem of detecting \acp{Ar}, \acp{LM}, and \acp{SDB}, the network contains three streams: the \ac{Ar} stream takes as input the \acp{EEG}, the \acp{EOG}, and the chin \ac{EMG} signals for a total of \(C_\text{Ar} = 5\) channels; the \ac{LM} stream receives the \(C_\text{LM}=2\) leg \ac{EMG} signals; and the \ac{SDB} stream receives the nasal pressure and the thoracoabdominal signals for a total of \(C_\text{SDB}=3\) channels.
An overview of the network architecture is shown graphically in~\cref{fig:architecture}.

\subsubsection{Stream specifics}
Each stream is comprised of two components.
First, a mixing module \( \varphi_\text{mix} : \R{C_{\ast} \times T} \rightarrow \R{C_{\ast} \times T} \) computes a non-linear mixing of the \(C\) channels using a set of \(C\) single-strided 1-dimensional filters \(\mathbf{w} \in \R{C \times C}\) and \ac{ReLU} activation~\cite{Nair2010}. 
Second, the output activations from \(\varphi_\text{mix}\) are used as input to a feature extraction module \(\varphi_\text{feat} : \R{C_\ast \times T} \rightarrow \R{f^{\prime} \times T^{\prime}}\), which transforms the input feature maps to a \(f^{\prime} \times T^{\prime}\) feature space with a temporal dimension reduced by a factor of \(\frac{T}{T^{\prime}}\).
The feature extraction module \(\varphi_\text{feat}\) is realized using \(k_\text{max}\) successive convolutions with an increasing number of filters \(f^{\prime} = f_0 2^{k-1}, \, k \in \llbracket k_\text{max} \rrbracket\), where \(f_0\) is a tunable base filter number.
Each feature map is normalized using batch normalization~\cite{Ioffe2015} with subsequent \ac{ReLU} activation.
\subsubsection{Feature fusion for sequential processing}
The output vectors from each stream is concatenated into a combined feature vector \(\mathbf{z}=\parentheses*{\mathbf{z}_\text{ar}, \mathbf{z}_\text{lm}, \mathbf{z}_\text{sdb}} \in \R{3f^{\prime} \times T^{\prime}}\).
We introduce sequential modeling of the feature vectors via a recurrent module \(\varphi_\text{rec} : \R{3f^{\prime} \times T^{\prime}} \rightarrow \R{2n_h \times T^{\prime}}\) realized with a \ac{bGRU}~\cite{Cho2014}.
The output of the \ac{bGRU} for timestep \textit{t} is a vector \( \mathbf{h}_{t} = \parentheses*{{\mathbf{h}_{t}^{\mathsf{f}}}, {\mathbf{h}_{t}^{\mathsf{b}}}} \in \R{2n_h}\) containing the concatenated outputs from the forward (\textsf{f}) and backward (\textsf{b}) directions.

\subsubsection{Additive attention}
We implemented a simple \textit{additive attention} mechanism~\cite{Bahdanau2015}, which computes \textit{context}-vectors \(\mathbf{c} \in \R{2n_h \times K}\) for each event class as the weighted sum of the feature vector outputs \(\mathbf{h}\in\R{2n_h \times T^{\prime}}\) from the \ac{bGRU}.

Formally, attention is computed as
\begin{IEEEeqnarray}{rCl}
    \mathbf{c} &=& \sum_{t}^{T^{\prime}} \mathbf{h}_{t} \boldsymbol{\alpha}_{t}^{\intercal}, 
\end{IEEEeqnarray}
where \(\mathbf{h}_t\) is the feature vector for time step \(t\), and \(\boldsymbol{\alpha}_t \in \R{K}\) is the attention weight computed as
\begin{equation}
    \boldsymbol{\alpha}_t = \frac{ \func{\exp}{\mathbf{W}_{a}\func{\tanh}{\mathbf{W}_\text{u}\mathbf{h}_t}}}{\sum_{\tau}^{T^{\prime}}\func{\exp}{\mathbf{W}_{a}\func{\tanh}{\mathbf{W}_\text{u}\mathbf{h}_\tau}}}.
\end{equation}
Here, \(\mathbf{W}_u \in \R{n_a \times 2n_h}\) and \(\mathbf{W}_a \in \R{K \times n_a}\) are learnable linear mappings of the feature vectors.

\subsubsection{Detection}
The final event classification and localization is handled by two modules, \(\psi_{\mathrm{clf}} : \R{2n_h \times K} \to \R{N_d \times K}\) and \( \psi_{\mathrm{loc}} : \R{2n_h \times K} \to \R{N_d \times 2} \), respectively.
The classification module \(\psi_\text{clf} : \mathbf{c} \mapsto \mathbf{p}\) outputs a tensor \(\mathbf{p} \in \brackets*{0, 1}^{N_d \times K}_{+} \) containing predicted event class probabilities for each default event window.
The localization module \(\psi_\text{loc} : \mathbf{c} \mapsto \mathbf{y}\) outputs a tensor \(\mathbf{y} \in \R{N_d \times 2}\) containing encoded relative onsets and durations for a detected event for each default event window.

\subsection{Training objective}
Similar to~\cite{Olesen2020b}, we optimized the network parameters according to a three-component loss function consisting of 
\begin{enumerate*}[before=\unskip{: }, itemjoin={{; }}, itemjoin*={{, and }}, label=\roman*)]
\item a localization loss \( \ell_{\mathrm{loc}} \)
\item a positive classification loss $\ell_{+}$
\item a negative classification loss $\ell_{-}$
\end{enumerate*}, 
such that the total loss $\ell$ was defined by
\begin{equation}\label{eq:loss}
    \ell = \ell_{\mathrm{loc}} + \ell_{+} + \ell_{-}.
\end{equation}
The localization loss $\ell_{\mathrm{loc}}$ was calculated using a Huber function
\begin{IEEEeqnarray}{rCl}
    \ell_{\mathrm{loc}} & = & \frac{1}{N_{+}} \, \sum_{\mathclap{i \in \pi_{+}}}\!{f_{H}^{(i)}}, \\
    \mathbf{f}_{H} & = &
    \begin{cases}
        0.5 \parentheses*{ \mathbf{y} - \mathbf{t} }^2, & \text{if } \abs*{ \mathbf{y} - \mathbf{t} } < 1, \\
        \abs*{ \mathbf{y} - \mathbf{t} } - 0.5, & \text{otherwise,}
    \end{cases}
\end{IEEEeqnarray}
where $i \in \pi_{+}$ yields indices of event windows with positive targets, \ie event windows matched to an arousal, \ac{LM} or \ac{SDB} target, and $N_{+}$ is the number of positive targets in the given data segment.

The positive classification loss component $\ell_{+}$ was calculated using a simple cross-entropy over the event windows matched to an arousal, \ac{LM}, or \ac{SDB} event:
\begin{IEEEeqnarray}{rCl}
    \ell_{+}&=&\frac{1}{N_{+}} \, \sum_{i \in \pi_{+}} \sum_{k \in \llbracket K \rrbracket} \pi^{(i)}_{k} \log p^{(i)}_{k}, \\ 
    p^{(i)}_{k}&=&\frac{\exp{s^{(i)}_{k}}}{\sum_{j} \exp s^{(i)}_{j}},{}
\end{IEEEeqnarray}
and $\pi^{(i)}_k$, $p^{(i)}_k$, and $s^{(i)}_k$ are the true class probability, predicted class probability, and logit score for the \textit{i}th event window containing a positive sample.

Similar to~\cite{Chambon2018b, Chambon2019}, the negative classification loss $\ell_{-}$ was calculated using a hard negative mining approach to balance the number of positive and negative samples in a data segment after matching default event windows to true events~\cite{Liu2016}.
Specifically, this is accomplished by calculating the probability for the negative class (no event) for each unmatched default event window, and then calculating the cross entropy loss using the \textit{Z} most probable samples.
In our experiments, we set the ratio of positive to negative samples as 1:3, such that the calculation of $\ell$ involves $Z=3$ times as many negative as positive samples.

We also explored a focal loss objective function for computing $\ell_{+}$ and $\ell_{-}$~\cite{Lin2020}, however, we found that this approach severely deteriorated the ability of the network to accurately detect \ac{LM} and \ac{SDB} events compared to using worst negative mining.

\subsection{Experimental setups}
\begin{table}[tb]
\footnotesize
    \centering
    \caption{MSED parameter settings.}
    \label{tab:parameters}
    \begin{tabular}{@{}LlC@{}}
        \toprule
        \textbf{Symbol} & \textbf{Description} & \textbf{Value} \\
        \midrule
        \ntrain & Number of train records & 1653 \\
        \neval & Number of eval records & 200 \\
        \ntest & Number of test records & 1000 \\
        C & Number of input channels & 10 \\
        \delta & Duration of default event windows & \SIlist[list-units=single,list-final-separator = {, }, list-pair-separator= {, }]{3;5;10}{\second} \\
        N_d & Number of default event windows & 208 \\
        f_s & Sampling rate & \SI{128}{\hertz} \\
        T & Segment length, samples & 15360 \\
        T^{\prime} & Reduced segment length, samples & 120 \\
        k_{\mathrm{max}} & Number of layers in \(\varphi_{\mathrm{feat}}\) & 7 \\
        f_0 & Base filter number & 4 \\
        f^{\prime} & Feature vector size & 512 \\
        n_h & Hidden units in \ac{bGRU} & 128 \\
        n_u & Hidden units in \(\varphi_{\mathrm{att}}\) & 128 \\
        K & Number of output classes & 4 \\
        (\beta_1, \beta_2) & Adam decay rates & (0.9, 0.999) \\
        \eta & Initial learning rate & \num{1e-3} \\
            & Learning rate decay factor & 0.1 \\
            & Learning rate decay step & 3 \text{ epochs} \\
            & Early stopping & 10 \text{ epochs} \\
        \bottomrule
    \end{tabular}
\end{table}

In our experiments, we optimized the training objective using adaptive moment estimation (Adam)~\cite{Kingma2015}, according to the loss function described in~\cref{eq:loss}. 
Exponential decay rates were fixed at \((\beta_1, \beta_2) = (0.9, 0.999)\), the learning rate at \(\eta = 10^{-3}\), and \(\epsilon = 10^{-8}\).
The learning rate was decayed in a step-wise manner by multiplying \( \eta \) with a factor of 0.1 after 3 consecutive epochs with no improvement in loss value on the validation dataset.
Similarly, we employed an early stopping scheme by monitoring the loss on the validation dataset and stopping the model training after 10 epochs of no improvement on \(\mathcal{D}_\textsc{eval}\).

We tested four types of models in two categories: the first is a default split-stream model as shown in~\cref{fig:architecture} with and without weight decay (splitstream, splitstream-wd).
The second is a variation of the split-stream model, but where the \(\psi_\text{clf}\) and \(\psi_\text{loc}\) modules are realized using depth-wise convolutions, such that each attention group is used only for that type of event.
The second category is also tested with and without weight decay (splitstream-dw, splitstream-dw-wd).

We benchmarked our proposed MSED model against DOSED by comparing overall performance on $\test$ after training on $\ntrain = 200 $ subject PSGs.
Each model was allowed 100 epochs of training, and the optimal models were selected based on lowest loss score on $\eval$ across epochs.
F1, precision, and recall scores were obtained by evaluating optimized models on $\test$.

Various model parameters are shown in~\cref{tab:parameters}.

\subsection{Performance evaluation}
Performance was quantified using precision, recall, and F1 scores.
Statistical significance in F1 score between groups was assessed with Kruskall-Wallis \textit{H}-tests.
The performance of joint vs. single-event detection models was tested with Wilcoxon signed rank tests for matched samples.
The relationships between true and predicted \ac{ArI}, \ac{AHI}, and \ac{LMI} were assessed using linear models and Pearson's \(r^2\).
Significance was set at \(\alpha=0.05\).
    
\section{Results and discussion}\label{sec:results}
\begin{figure*}[t]
    \centering
    \includegraphics[width=\textwidth]{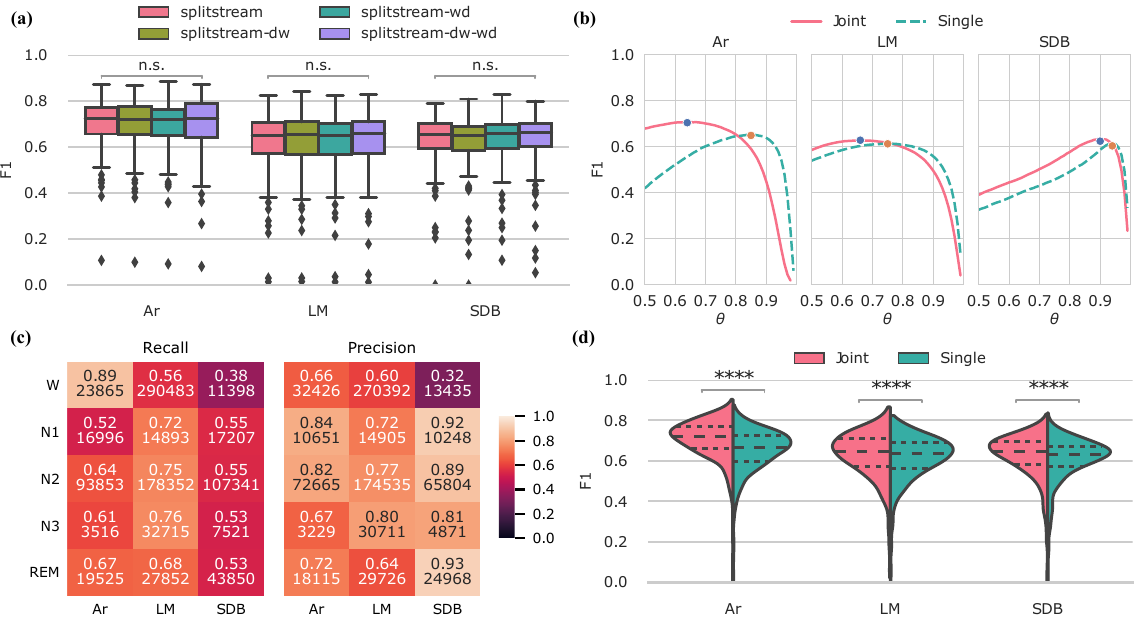}
    \caption{Optimizing MSED architecture and model evaluation. %
    \textbf{(a)} No significant differences in F1 were found between model architectures. %
    \textbf{(b)} Optimizing F1 performance on \eval as a function of detection threshold \(\theta\) for joint (solid) and single-event (dashed) detection models. %
    Blue and orange dots correspond to optimal \(\theta\)/F1 pair. %
    \textbf{(c)} Analysing event recall and precision categorized by sleep stage origin on \test data. %
    E.g. out of 19525 \ac{Ar} events manually scored in \ac{REM}, the optimal model correctly identifies 0.67\%, while out of 10248 \ac{SDB} events predicted in \ac{N1}, 0.92\% of these are also manually scored. %
    \textbf{(d)} Evaluating optimized joint and single-event detection models on \test given the respective detection threshold \(\theta\) from \textbf{(b)}. 
    For all three event types, the joint detection model outperforms the single-event models based on F1.
    Dashed lines in the violin plot interior show the associated \nth{25}, \nth{50}, and \nth{75} percentiles. %
    \(^{****}\): \(p < 0.001\). %
    \describe{W}; %
    \describe{N1}; %
    \describe{N2}; %
    \describe{N3}; %
    \describe{REM}; %
    \describe{Ar}; %
    \describe{LM}; %
    \describe{SDB}.%
    }
    \label{fig:results}
\end{figure*}
\begin{table*}[htb]
\footnotesize
    \centering
    \begin{threeparttable}
    \caption{Temporal difference metrics across event types and PSGs for joint and single prediction models. 
    Positive values of onset/offset correspond to delayed predictions relative to the matched true event, while positive duration values correspond to an overestimation of event duration. 
    Metrics are shown as mean ± standard deviation with 95\% confidence interval in brackets across all events for all PSG recordings.
    }
    \label{tab:test_temporal_metrics}
    \begin{tabularx}{\textwidth}{@{}llYYY@{}}
        \toprule
         & & \textbf{\textDelta duration} & \textbf{\textDelta onset} & \textbf{\textDelta offset} \\
        \midrule
        \ac{Ar}  & Joint  & \ci{4.81}{2.92}{1.07}{10.77} & \ci{-1.37}{1.39}{-3.99}{0.33} & \ci{3.44}{2.24}{0.56}{7.41} \\
                 & Single & \ci{6.42}{3.70}{1.37}{12.99} & \ci{-1.94}{1.93}{-5.33}{0.50} & \ci{4.49}{2.69}{0.85}{9.21} \\ \midrule
        \ac{LM}  & Joint  & \ci{-0.46}{0.36}{-1.05}{0.07} & \ci{0.24}{0.22}{-0.04}{0.56} & \ci{-0.23}{0.20}{-0.55}{0.05} \\
                 & Single & \ci{-0.51}{0.39}{-1.16}{0.01} & \ci{0.26}{0.23}{-0.02}{0.61} & \ci{-0.25}{0.21}{-0.61}{0.06} \\ \midrule
        \ac{SDB} & Joint  & \ci{20.41}{13.36}{3.50}{43.77} & \ci{-9.39}{6.74}{-21.24}{-0.97} & \ci{11.02}{7.67}{1.14}{23.70} \\
                 & Single & \ci{34.95}{16.63}{10.81}{63.18} & \ci{-16.00}{8.80}{-32.35}{-3.65} & \ci{18.95}{9.51}{5.44}{35.33} \\
        \bottomrule
    \end{tabularx}
    \begin{tablenotes}
    \item %
    \describe{Ar}; %
    \describe{LM}; %
    \describe{SDB}.
    \end{tablenotes}
    \end{threeparttable}
\end{table*}

\subsection{Model architecture evaluation}
We found no significant differences in F1 performance for either \ac{Ar} (Kruskal-Wallis \(H = 0.96, \, p = 0.81\)), \ac{LM} (\( H = 0.23, \, p = 0.97 \)), or \ac{SDB} detection (\( H=2.84, \, p = 0.42 \)), when evaluating the model architectures on \eval (see~\cref{fig:results}a).
Subsequent results are thus reported based on the default splitstream architecture.

\subsection{Optimizing threshold for joint vs. single event detection}
\begin{table}[htb]
\footnotesize
    \centering
    \begin{threeparttable}
    \caption{Performance scores for optimized models evaluated on \test. %
             Metrics are shown aggregated across \acp{PSG} as mean ± standard deviation. %
             Overall metrics are computed by averaging scores for each subject \ac{PSG}.}
    \label{tab:test_performance}
    \begin{tabularx}{\columnwidth}{@{}llYY@{}}
        \toprule
         & & \textbf{Joint} & \textbf{Single} \\
        \midrule
         \ac{Ar} &     F1 & \textbf{\plusminus{0.704}{0.106}} & \plusminus{0.649}{0.113} \\
                 &     Pr & \plusminus{0.759}{0.114} & \plusminus{0.777}{0.107} \\
                 &     Re & \plusminus{0.672}{0.125} & \plusminus{0.571}{0.127} \\ \midrule
         \ac{LM} &     F1 & \plusminus{0.628}{0.123} & \plusminus{0.613}{0.116} \\
                 &     Pr & \plusminus{0.650}{0.169} & \plusminus{0.661}{0.166} \\
                 &     Re & \plusminus{0.647}{0.120} & \plusminus{0.607}{0.116} \\ \midrule
        \ac{SDB} &     F1 & \plusminus{0.624}{0.115} & \plusminus{0.604}{0.118} \\
                 &     Pr & \plusminus{0.817}{0.142} & \plusminus{0.835}{0.136} \\
                 &     Re & \plusminus{0.526}{0.146} & \plusminus{0.492}{0.145} \\ \midrule
         Overall &     F1 & \plusminus{0.652}{0.121} & \plusminus{0.622}{0.117} \\
                 &     Pr & \plusminus{0.742}{0.159} & \plusminus{0.758}{0.157} \\
                 &     Re & \plusminus{0.615}{0.146} & \plusminus{0.556}{0.138} \\
        \bottomrule
    \end{tabularx}
    \begin{tablenotes}
    \item %
    \describe{Ar}; %
    \describe{LM}; %
    \describe{SDB}; %
    Pr: precision; %
    Re: recall.
    \end{tablenotes}
    \end{threeparttable}
\end{table}
For each event type, we evaluated the F1 score as a function of classification threshold \(\theta\) on \eval for both the joint detection model as well as the single-event models.
It can be observed in~\cref{fig:results}b that for all three events, the joint detection model achieves higher F1 score, although the increase is not as large for \ac{LM} and \ac{SDB} detection.
This was also observed when evaluating the joint and single detection models with optimized thresholds on \test for both \ac{Ar} (Wilcoxon \(W = 30440, \, p < 0.001\)), \ac{LM} (\(W = 101103, \, p < 0.001\)), and \ac{SDB} detection (\(W = 125461, \, p < 0.001\)), see~\cref{fig:results}d. 
Precision, recall and F1 scores for optimized models evaluated on \text are shown in~\cref{tab:test_performance}.
These findings provide evidence that the presence of certain event types can modulate the detection of other event types, and that this can be modeled using automatic methods.
This is in line with what previous studies have found \eg on event-by-event scoring agreement in arousals, which improved significantly from \SIrange{58.7}{90.5}{\percent}, when including respiratory signals in the analysis~\cite{Thomas2003}.
\subsection{Comparison with state-of-the-art multi-event detection}
F1, recall, and precision scores for optimized DOSED and MSED models evaluated on \(\test\) are shown in~\cref{tab:benchmark}.
We observed an overall MSED F1 score of 0.634 ± 0.124 against an overall DOSED F1 score of 0.596 ± 0.140; and overall F1 scores for \ac{Ar}, \ac{LM}, and \ac{SDB} of 0.677 ± 0.107, 0.599 ± 0.127, and 0.626 ± 0.125 for MSED, and 0.668 ± 0.115, 0.619 ± 0.125, and 0.503 ± 0.123 for DOSED.
Factoring in the overall reduction in model size from 385,489,502 to 9,435,606 parameters, these results show the advantage of MSED compared to DOSED on the same dataset.

Comparing with existing single-event arousal detection models, MSED does not perform on the level of previous state-of-the-art proposed by~\citeauthor{Brink-Kjaer2020}~\cite{Brink-Kjaer2020}. 
Here, an EEG+EOG+EMG combination model for combined sleep-wake classification and arousal detection yielded an F1 score of 0.76, although this was reported on a much smaller dataset.
Similarly, in the work by~\citeauthor{Carvelli2020}, a model combining two leg EMG channels achieved an impressive F1 score of 0.77~\cite{Carvelli2020}, although this was also reported in a much smaller dataset.
We did not perform in-depth ablations in this study, so it is possible that the MSED performance could be higher given sufficient fine-tuning.
However, it is also worth noting that both of these models apply post-processing of the model outputs, most notably the removal of arousals and leg movements scored in wake, which is not performed in the current work, and fusion of events within certain manually-set thresholds.
It is possible that this approach introduces a negative bias in our proposed model, since it is evident from~\cref{fig:results}c that the precision for all scored events is lower in~\ac{W} than in other sleep stages.
In this work we wished the predictions to be orthogonal to manual sleep scoring, but future work should consider adding an automatic sleep scoring module to the model architecture.

While literature on sleep apnea detection is extensive, it is difficult to compare directly to the proposed approach, because the majority of studies either focus on obstructive apnea alone, do not report F1, precision, or recall scores, or only focus on the prediction of \ac{AHI} alone~\cite{Mendonca2019}. 

However, one recent study compared the event-by-event detection performance against a concensus score of five technicians.
They reported an average human performance quantified by F1 of 0.55, and an F1 score from the automatic method of 0.57~\cite{Thorey2019}
Similarly, \citeauthor{Nassi2022} recently proposed their WaveNet model for precisely annotating \ac{SDB} events in 1 s bins.
Although their model also included post-processing of the bins, they obtained a mean F1 score across events of 0.406.
We therefore see a marked improvement from the state of the art event detectors compared to MSED.
    
These results also indicate the massive research potential in terms of other ways to assess \ac{SDB}; apart from AHI, which is an average across the entire night, researchers and clinicians could potentially benefit from taking a more fine-grained approach. 
As illustrated by~\citeauthor{Chen2022}, patients with the exact same AHI can exhibit wildly different activity patterns (breathing disturbances)~\cite{Chen2022}, yet this is unaccounted for in state-of-the-art apnea detection models, as the majority of these are epoch-based~\cite{Mendonca2019}. 
\citeauthor{Chen2022} proposes "instantaneous AHI" as a solution to this problem, although their results were based on human annotation and not automatic detection as proposed here.
The integration of these two methods would be an interesting approach for future research.

In addition, future work should explore novel methods for object detection in the computer vision literature.
Most notably, the use of default event windows impose certain restrictions on the temporal scale of detected events, and this could be eliminated by using a Transformer-based detection model, such as the one found in the Detection Transformer~\cite{Carion2020}.
Here, predictions are generated using a number of object queries, which are independent from the default event windows and thus not restricting the temporal scale of the events.

\begin{table}[tb]
\footnotesize
    \centering
    \begin{threeparttable}
    \caption{Comparing model performance against DOSED on a reduced \(\train\).
             Metrics are shown aggregated across \acp{PSG} as mean ± standard deviation. 
             Overall metrics are computed by averaging scores for each subject \ac{PSG}.}
    \label{tab:benchmark}
    \begin{tabularx}{\columnwidth}{@{}llYY@{}}
        \toprule
         & & \textbf{DOSED} & \textbf{MSED} \\
        \midrule
        \ac{Ar} & F1 & \plusminus{0.668}{0.115} & \plusminus{0.677}{0.107} \\
                & Pr & \plusminus{0.711}{0.120} & \plusminus{0.688}{0.118} \\
                & Re & \plusminus{0.651}{0.141} & \plusminus{0.686}{0.131} \\ \midrule
        \ac{LM} & F1 & \plusminus{0.619}{0.125} & \plusminus{0.599}{0.127} \\
                & PR & \plusminus{0.634}{0.169} & \plusminus{0.655}{0.174} \\
                & Re & \plusminus{0.644}{0.120} & \plusminus{0.587}{0.133} \\ \midrule
        \ac{SDB}& F1 & \plusminus{0.503}{0.123} & \plusminus{0.626}{0.125} \\
                & Pr & \plusminus{0.706}{0.173} & \plusminus{0.793}{0.146} \\
                & Re & \plusminus{0.412}{0.133} & \plusminus{0.542}{0.157} \\ \midrule
        Overall & F1 & \plusminus{0.596}{0.140} & \plusminus{0.634}{0.124} \\
                & Pr & \plusminus{0.684}{0.160} & \plusminus{0.712}{0.159} \\
                & Re & \plusminus{0.568}{0.172} & \plusminus{0.605}{0.153} \\ \midrule
        No. parameters & & 385,489,502          & 9,435,606                \\
        \bottomrule
    \end{tabularx}
    \begin{tablenotes}
    \item %
    \describe{Ar}; %
    \describe{LM}; %
    \describe{SDB}; %
    Pr: precision; %
    Re: recall.
    \end{tablenotes}
    \end{threeparttable}
\end{table}

\subsection{Correlation between experts and model}
\begin{figure}[tb]
    \centering
    \includegraphics[width=\linewidth]{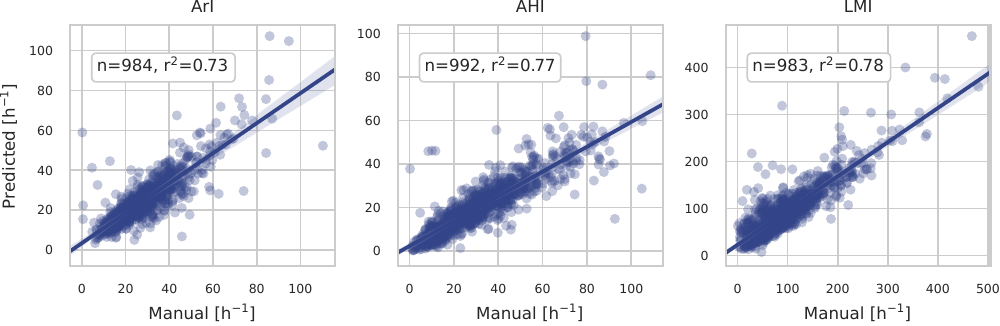}
    \caption[Correlation between manually scored and predicted indices]{Pearson correlation plots for each event type index between true and predicted values. The linear relationship is indicated with solid blue with 95\% confidence intervals in light blue.
    \describe{ArI}; %
    \describe{AHI}; %
    \describe{LMI}.}
    \label{fig:event-detection:paper-vi:correlation_plot}
\end{figure}

For each event type, we computed the correlation coefficient between predicted and expert-annotated ArI, AHI, and LMI, which is shown in~\cref{fig:event-detection:paper-vi:correlation_plot}.
We found a large positive correlation between true and predicted values for \ac{ArI} (\(r^2=0.73\), \(p < 0.001\)), \ac{AHI} (\(r^2=0.77\), \(p < 0.001\)), and \ac{LMI} (\(r^2=0.78\), \(p < 0.001\)).

A similar study by~\citeauthor{Biswal2018} using an automatic method for automatic detection of \ac{SDB} and \ac{LM} events found similar or higher correlations between automatic and manual scoring (\(r^2_{\mathrm{SDB}} = 0.85\), and \(r^2_{\mathrm{LM}} = 0.79\), respectively), although their findings were based on almost 5 times as much data~\cite{Biswal2018}.
Furthermore, obstructive, central, mixed and hypopneas with an associated 4\% desaturation were lumped together into a single \textit{apnea} class, which may have introduced unwanted bias towards obstructive apneas and hypopneas in their findings, since these are in general more prevalent than central and mixed apneas.

\subsection{Temporal difference metrics}
We compared the temporal precision between manual and MSED event scoring by looking at the errors in onset (\(\Delta \text{onset}\)), offsets (\(\Delta \text{fffset}\)), and durations (\(\Delta \text{duration}\)) calculated as
\begin{align}
    \Delta\,\text{onset} &= \text{onset}_{\text{MSED}} - \text{onset}_{\text{manual}} \\
    \Delta\,\text{offset} &= \text{offset}_{\text{MSED}} - \text{offset}_{\text{manual}} \\
    \Delta\,\text{duration} &= \text{duration}_{\text{MSED}} - \text{duration}_{\text{manual}}
\end{align}
so that positive values of \(\Delta\,\text{onset}, \Delta\,\text{offset}\) corresponds to a positive shift to the right (delayed prediction), and positive values of \(\Delta\,\text{duration}\) meaning an overestimation of the event duration compared to manual scoring.

Described in~\cref{tab:test_temporal_metrics}, the model overestimates the duration of \ac{Ar} events by a couple of seconds, which is caused by an earlier prediction of onset and delayed prediction of termination.
For \ac{LM} events, the model underestimates the duration by about half a second on average, which is due to earlier prediction of termination.
For \ac{SDB} events, the model overestimates the duration by about 25 seconds on average, which is caused by an earlier prediction of onset and delayed prediction of termination.
These errors in predicted duration reflect the temporal characteristics of these events; \acp{LM} are shorter events (between \SIrange{0.5}{10}{\second} per definition), and it is thus unlikely to be overestimated by several seconds, while \acp{SDB} are longer events by one to two orders of magnitude, which also increases the size of the errors. 
Arousals are intermediate in length compared to \acp{LM} and \acp{SDB}, which is reflected in the error distributions.
    
\section{Conclusion}
We have presented a novel method for detecting short and long events present in polysomnogram recordings based on deep neural networks.
Our method was able to distinguish between arousals, limb movements, and sleep-disordered breathing events with F1 scores of 0.70, 0.63, and 0.62, respectively, and we furthermore found that jointly optimizing a model for all three events performed better than the respective models optimized for each specific event type.

We benchmarked our algorithm against previous state-of-the-art and report an overall increase in F1 score from 0.60 to 0.63 despite a 97.5\% reduction in memory footprint.

Furthermore, clinically relevant derivatives (\ac{ArI}, \ac{AHI}, \ac{LMI}) showed a high positive correlation with manually computed values indicating a high degree of agreement between our model and experts.

Future work should incorporate ides from the object detection in computer vision literature and investigate more complex models with increased flexibility towards adding prediction capabilities for additional event types.
Additionally, the low precision across all events observed during wakefulness could be remedied by incorporating an automatic sleep stage classification model which also merits further investigation.

\appendices

\section*{Acknowledgment}
The National Heart, Lung, and Blood Institute provided funding for the ancillary MrOS Sleep Study, "Outcomes of Sleep Disorders in Older Men," under the following grant numbers: R01 HL071194, R01 HL070848, R01 HL070847, R01 HL070842, R01 HL070841, R01 HL070837, R01 HL070838, and R01 HL070839. The National Sleep Research Resource was supported by the National Heart, Lung, and Blood Institute (R24 HL114473, 75N92019R002).

Some of the computing for this project was performed on the Sherlock cluster. 
We would like to thank Stanford University and the Stanford Research Computing Center for providing computational resources and support that contributed to these research results.

The authors would like to thank Andreas Brink-Kjær, Rasmus Malik Thaarup Høegh, Anders Stevnhoved Olsen, Mads Olsen, and Laura Rose for their valuable comments and feedback in preparing this manuscript.

\printbibliography

\end{document}